\documentclass[11pt]{article}

\usepackage{acl}

 \usepackage{microtype}

\usepackage{microtype}
\usepackage{graphicx}
\usepackage{subcaption}
\usepackage{multirow}
\usepackage{booktabs} % for professional tables
\usepackage{tabularx}   % provides tabularx environment
\usepackage{array}      % provides >{\raggedright\arraybackslash} etc.
\usepackage{float}      % provides [H] specifier
\usepackage{placeins}   % provides \FloatBarrier
\usepackage{hyperref}

\newcommand{\Linearize}{\mathrm{Linearize}}

\usepackage{amsmath}
\usepackage{amssymb}
\usepackage{mathtools}
\usepackage{amsthm}

\usepackage[capitalize,noabbrev]{cleveref}

\theoremstyle{plain}

\theoremstyle{definition}

\theoremstyle{remark}

\usepackage[english,bidi=default]{babel} % English as the main language.
\babelprovide[import]{hindi}

\babelprovide[import]{arabic}

\title{When Evidence Shapes Collaboration: Knowledge-Conditioned Topology Generation for Multi-Agent Systems}

\author{
Yangxiao Jiang\thanks{Equal contribution.} \quad
Jiarun Fan\footnotemark[1] \quad
Mingcong Xu\footnotemark[1] \quad
Yanxi Guo \quad
Jiwen Feng \\
Shanqing Xu \quad
Mengchen Qian \quad
Wei Chen \quad
Xiaojin Zhang\thanks{\mbox{Corresponding author. \texttt{xiaojinzhang@hust.edu.cn}}} \\
\\
Huazhong University of Science and Technology
}

\begin{document}

\maketitle
\begin{abstract}
Multi-Agent Systems (MAS) have recently moved from static workflows toward dynamically generated collaboration topologies. However, existing topology generation methods rely primarily on the parametric knowledge of large language models, with external search or retrieval used only as a reactive tool rather than an explicit determinant of collaboration structure. This leads to structure–knowledge misalignment, where systems exhibit redundant interactions or insufficient verification in knowledge-intensive tasks. We propose K-GAT (Knowledge-Guided Agent Topology Generator), a neuro-symbolic framework that formulates collaboration topology design as a knowledge-conditioned structure learning problem, integrating external evidence directly into autoregressive graph generation. Extensive experiments on knowledge-intensive benchmarks demonstrate K-GAT's efficiency and effectiveness: notably on the expert-level GPQA dataset, K-GAT outperforms the LLM-Debate baseline by a substantial margin of +15.7\% in accuracy, while consuming less than half the computational tokens.
\end{abstract}

\begin{figure*}[t]
    \centering
    % --- Subfigure (a): Over-planning ---
    \begin{subfigure}[b]{0.32\textwidth}
        \centering
        % 替换为您第一张图的文件名
        \includegraphics[width=\textwidth]{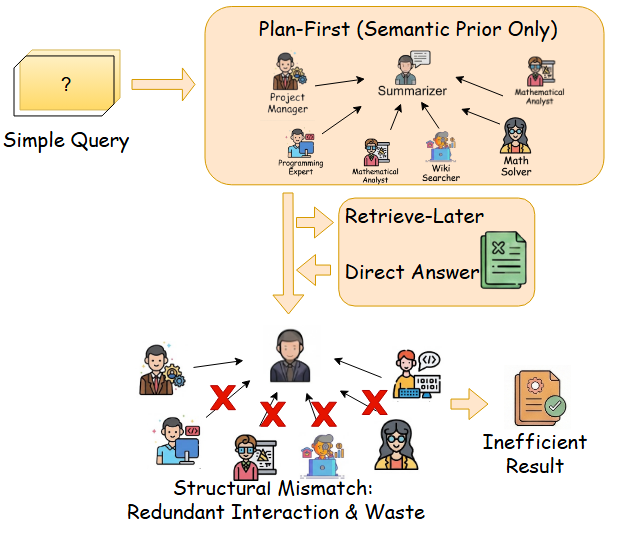}
        \caption{Plan-First (Over-planning)}
        \label{fig:mismatch_over}
    \end{subfigure}
    \hfill
    % --- Subfigure (b): Under-planning ---
    \begin{subfigure}[b]{0.32\textwidth}
        \centering
        % 替换为您第二张图的文件名
        \includegraphics[width=\textwidth]{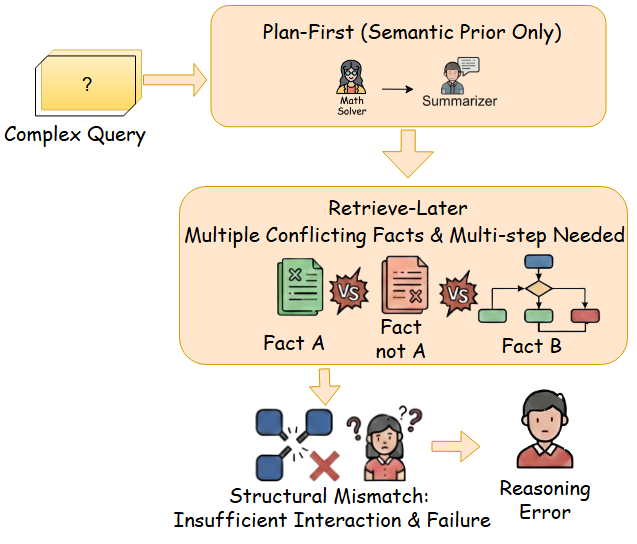}
        \caption{Plan-First (Under-planning)}
        \label{fig:mismatch_under}
    \end{subfigure}
    \hfill
    % --- Subfigure (c): K-GAT (Ours) ---
    \begin{subfigure}[b]{0.32\textwidth}
        \centering
        % 替换为您第三张图的文件名
        \includegraphics[width=\textwidth]{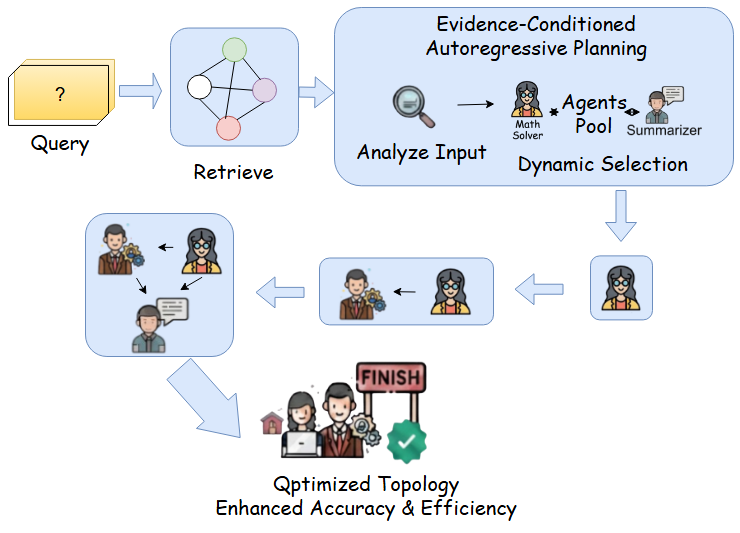}
        \caption{Evidence-First (K-GAT)}
        \label{fig:kgat_solution}
    \end{subfigure}
    
    \caption{Comparison of Planning Paradigms. 
(a) Over-planning: Semantically complex queries may retrieve simple and highly consistent evidence, yet semantic planning still generates unnecessarily sophisticated workflows. 
(b) Under-planning: Semantically simple queries may retrieve sparse or conflicting evidence, while semantic planning fails to allocate sufficient verification and collaboration. 
(c) K-GAT (Ours): An evidence-first approach that dynamically conditions the collaboration topology on the retrieved evidence.}
    \label{fig:comparison}
\end{figure*}

\section{Introduction}

Recent advances in large language models (LLMs) have transformed intelligent agents from isolated reasoning systems into collaborative problem-solving frameworks. Early agentic paradigms primarily relied on single-agent reasoning strategies, such as ReAct~\cite{yao2022react}, where retrieval and action execution were interleaved within a unified reasoning trajectory. While effective for relatively simple tasks, such approaches often struggle with complex reasoning scenarios requiring decomposition, verification, and parallel exploration. To address these limitations, recent studies have increasingly shifted toward Multi-Agent Systems (MAS), where multiple specialized agents collaboratively solve tasks through structured communication and coordination~\cite{li2023camel,wu2023autogen,liang2023encouraging,hong2024metagpt}. Early MAS frameworks typically relied on manually designed static collaboration topologies, such as fixed chains, debate pipelines, which provide stable execution patterns but lack adaptability to diverse reasoning requirements. More recently, dynamic topology generation methods~\cite{zhang2024g,li2025assemble} have emerged, enabling orchestration modules to automatically synthesize collaboration structures conditioned on task semantics. Such approaches offer a promising balance between the flexibility of general-purpose LLMs and the structured decomposition capabilities of collaborative reasoning.

Despite their flexibility, existing dynamic MAS frameworks still follow a ``Plan-First, Retrieve-Later'' paradigm: the collaboration topology is generated solely from query semantics before external retrieval is performed. However, semantic complexity does not necessarily reflect the actual evidence requirements of a task. Queries with seemingly complex semantics may require only minimal evidence aggregation, while tasks involving sparse or conflicting evidence may demand substantially more sophisticated coordination. As a result, the generated workflow can become misaligned with the retrieved evidence, leading to what we term \textit{Structural Mismatch}. As illustrated in Figure~\ref{fig:comparison}, this mismatch manifests in two representative failure modes. In some cases, the planner hallucinates unnecessarily sophisticated debate workflows even when the retrieved evidence is already sufficient and highly consistent, leading to redundant interactions and wasted computation (over-planning, Figure~\ref{fig:mismatch_over}). Conversely, when evidence is sparse, ambiguous, or conflicting, the planner may impose oversimplified sequential structures that fail to support sufficient verification or exploration (under-planning, Figure~\ref{fig:mismatch_under}). 

To address this issue, we propose \textit{Evidence-First Collaboration}, a paradigm where external evidence directly conditions workflow generation. Instead of treating retrieval as a passive component executed after planning, K-GAT (Knowledge-Guided Agent Topology Generator) generates collaboration topologies conditioned on retrieved evidence and provenance information. As shown in Figure~\ref{fig:kgat_solution}, K-GAT autoregressively constructs collaboration graphs by adaptively determining agent instantiation and communication routing under the retrieved evidence context. By grounding structural planning in external evidence, K-GAT improves reasoning reliability while reducing unnecessary collaboration overhead.

In summary, we identify Structural Mismatch as a key limitation of existing dynamic MAS frameworks and propose an evidence-conditioned workflow generation paradigm for multi-agent reasoning. Building upon this insight, K-GAT learns to generate adaptive collaboration topologies grounded in retrieved external evidence. Extensive experiments on knowledge-intensive reasoning benchmarks demonstrate that K-GAT consistently improves reasoning accuracy and workflow adaptability across diverse evidence conditions.

\section{Related Works}

Recent advances in LLM-based agents have shifted intelligent systems from single-agent reasoning toward collaborative multi-agent workflows. Early agentic frameworks, such as ReAct~\cite{yao2022react}, interleaved reasoning and tool use within a unified reasoning trajectory, while subsequent MAS introduced structured collaboration through predefined communication protocols and role assignments~\cite{li2023camel,hong2024metagpt,wu2023autogen,liang2023encouraging,du2024improving}. To overcome the rigidity of manually designed workflows, recent studies have explored automatic workflow and topology generation, enabling orchestration modules to dynamically synthesize agent roles and communication structures conditioned on task semantics~\cite{zhang2024g,li2025assemble,zhang2025aflow,zhou2025multi,zhang2025maas,shi2025flowxpert}.  Despite these advances, existing methods still primarily generate collaboration workflows from query semantics before external retrieval is performed. In contrast, our work argues that workflow generation should instead be conditioned on the epistemic properties of retrieved evidence, enabling collaboration structures to directly adapt to the underlying evidence landscape.

\section{Methodology}

\subsection{Evidence-Driven Topology Generation}
\label{sec:method}

We study knowledge-intensive reasoning tasks where answering a query
$\mathcal{Q}$ requires grounding on external evidence rather than relying solely on parametric knowledge.
Given a query--answer dataset
$\mathcal{D}=\{(\mathcal{Q}_i,Y_i)\}_{i=1}^{N}$,
a Multi-Agent System (MAS) solves $\mathcal{Q}$ by instantiating multiple LLM-based agents under a collaboration topology.
Each agent role $r \in \mathcal{R}_{\mathcal{T}}$
(e.g., Planner, Reasoner, Verifier, Answerer)
is associated with a task-specific system prompt
$\mathcal{S}_{sys}^{(r)}$.

Unlike existing dynamic MAS frameworks that generate workflows solely from query semantics,
K-GAT follows an \textit{Evidence-First} paradigm:
external retrieval is performed before topology generation,
and the retrieved evidence directly conditions both
(i) which agent roles should be instantiated and
(ii) how information should flow among agents.
The key intuition is that collaboration complexity should adapt to the retrieved evidence rather than relying purely on semantic priors from the query itself.

To construct the evidence context,
we retrieve a set of relevant evidence units
$\mathcal{T}_{sub}$ from an external structured knowledge source together with their provenance information $\Pi$.
Given a query $\mathcal{Q}$,
dense similarity scores are computed between the query embedding and candidate evidence units:
\begin{equation}
s(\tau;\mathcal{Q})=\mathrm{sim}(\mathbf{x}_{\tau},\mathbf{q}),
\end{equation}
where $\mathbf{q}$ denotes the embedding of $\mathcal{Q}$
and $\mathbf{x}_{\tau}$ denotes the embedding of evidence unit $\tau$.
The top-$K$ retrieved units form the evidence context:
\begin{equation}
\mathcal{C}_{\mathcal{K}}=(\mathcal{T}_{sub}, \Pi).
\end{equation}

We represent the collaboration workflow as a directed acyclic graph (DAG)
$\mathcal{G}=(\mathcal{V},\mathcal{E})$,
where each node corresponds to an instantiated agent role
and each directed edge indicates message passing between agents.
To encourage efficient collaboration, we define the structural cost:
\begin{equation}
C(\mathcal{G})
=
\lambda_V |\mathcal{V}|
+
\lambda_E |\mathcal{E}|,
\end{equation}
where smaller graphs correspond to lower communication and computation overhead.

Conditioned on the retrieved evidence context,
K-GAT incrementally generates the collaboration topology in an autoregressive manner:
\begin{equation*}
\begin{split}
P_{\theta}(\mathcal{G}\mid \mathcal{Q},\mathcal{C}{\mathcal{K}})
&= \prod{t} P_{\theta}(v_t \mid \mathcal{G}{<t},\mathcal{Q},\mathcal{C}{\mathcal{K}}) \
\\& \times \prod_{j<t} P_{\theta}(e_{j,t}\mid v_j,v_t),
\end{split}
\end{equation*}
\vspace{-1em}

where $v_t$ denotes the agent role instantiated at step $t$
and $e_{j,t}$ denotes the communication edge between agents.
During generation, structural constraints including acyclicity and bounded indegree are enforced to guarantee executable collaboration workflows.

\subsection{Curriculum Learning Framework}
\label{sec:training}

\begin{figure*}[t]
    \centering
    \includegraphics[width=\textwidth]{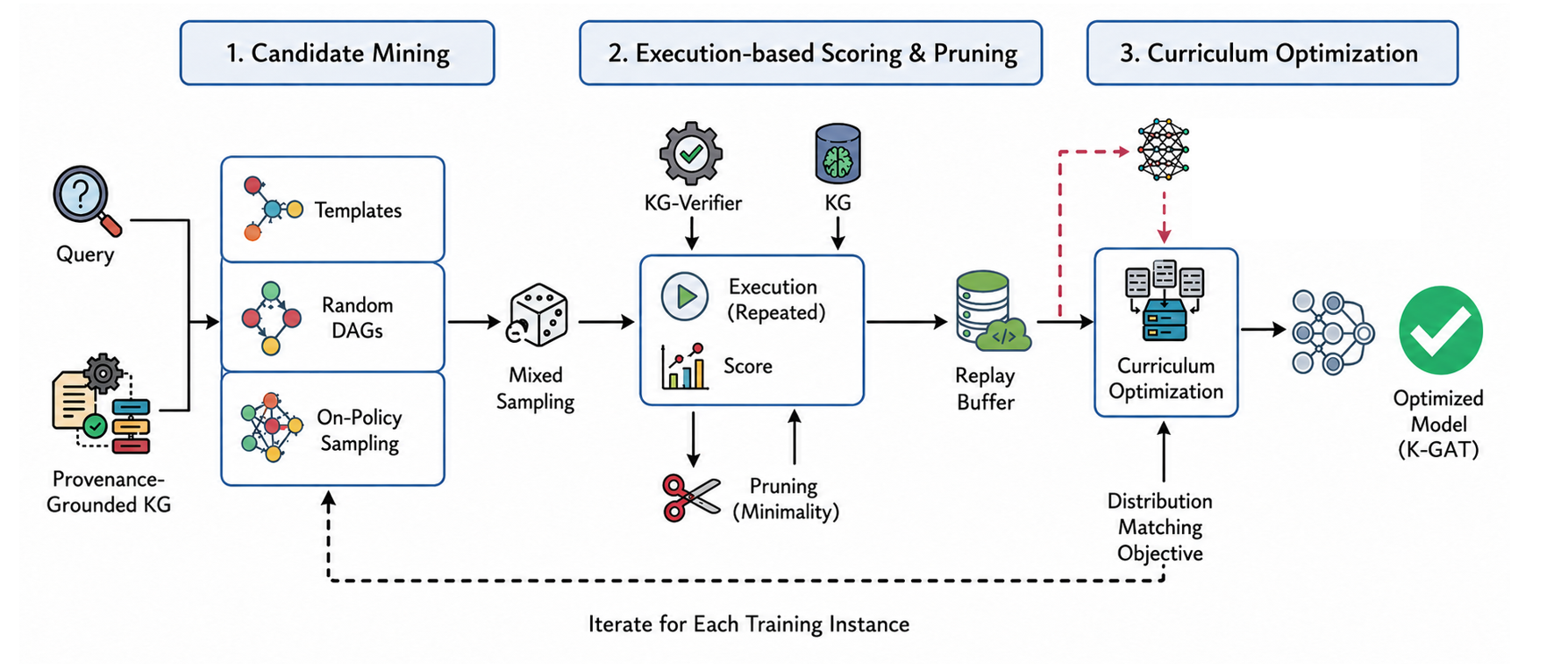}
    \vspace{-3mm}
    \caption{\textbf{Overview of the K-GAT Curriculum Training Framework.}
    The training pipeline iterates through three stages:
    (1) \textbf{Candidate Mining} (Left), where diverse topology candidates are sourced from canonical motifs, random DAGs, and on-policy sampling conditioned on the provenance-grounded KG;
    (2) \textbf{Execution-based Scoring \& Pruning} (Middle), where candidates are evaluated via actual execution and pruned to minimal sufficient structures ($\mathcal{G}^*$) to form a replay buffer; and
    (3) \textbf{Curriculum Optimization} (Right), where the generator is trained to learn these structures using a Distribution-matching objective.}
    \label{fig:training_framework}
    \vspace{-6mm}
\end{figure*}

Training K-GAT requires structural supervision, but standard knowledge-intensive QA datasets only provide query--answer pairs rather than gold collaboration workflows.
Given a training instance $(\mathcal{Q},Y)$, we first retrieve the top-$K$ evidence units and form the evidence context
$\mathcal{C}_{\mathcal{K}}=(\mathcal{T}_{sub},\Pi)$,
where $\mathcal{T}_{sub}$ denotes the retrieved evidence units and $\Pi$ denotes their provenance information.
Instead of assigning a single gold topology to each query, we construct a candidate set of possible workflows, evaluate them through execution, and train the generator to learn a conditional distribution over high-quality topologies for each $(\mathcal{Q},\mathcal{C}_{\mathcal{K}})$.

For each training instance, candidate topologies are sampled from three sources:
(i) predefined workflow templates $\mathcal{P}_{temp}$,
including chain, star, tree, and verifier-involved structures;
(ii) randomly sampled feasible DAGs $\mathcal{P}_{rand}$ satisfying the same structural constraints;
and (iii) on-policy samples generated from the current topology generator.
Candidate graphs are sampled from the mixed distribution:
\begin{equation}
\begin{split}
\mathcal{G} &\sim (1-\pi), \mathrm{Unif}(\mathcal{P}{temp}\cup\mathcal{P}{rand}) \
\\&\quad + \pi, P_\theta(\cdot\mid \mathcal{Q},\mathcal{C}_{\mathcal{K}}),
\end{split}
\end{equation}
where $\pi\in[0,1]$ balances exploration and exploitation.
Smaller $\pi$ values rely more on template and random candidates,
while larger values increase the influence of the current generator.

Each sampled topology $\mathcal{G}$ is executed under the same query and evidence context.
Because LLM-based execution is stochastic, we repeat execution $R$ times and estimate the empirical success rate:
\begin{equation}
\hat{p}_{succ}(\mathcal{G})
=
\frac{1}{R}
\sum_{r=1}^{R}
\mathbb{I}_{succ}(\hat{Y}^{(r)},Y),
\end{equation}
where $\hat{Y}^{(r)}$ denotes the prediction from the $r$-th execution and
$\mathbb{I}_{succ}(\hat{Y}^{(r)},Y)$ indicates whether the prediction is correct under the task evaluator.

To discourage unnecessarily large workflows, we define the structural cost:
\begin{equation}
C(\mathcal{G})
=
\lambda_V|\mathcal{V}|
+
\lambda_E|\mathcal{E}|,
\end{equation}
where $|\mathcal{V}|$ and $|\mathcal{E}|$ denote the numbers of agents and communication edges,
and $\lambda_V,\lambda_E$ control the relative cost of adding nodes and edges.
The final candidate score is:
\begin{equation}
\mathrm{Score}(\mathcal{G})
=
\hat{p}_{succ}(\mathcal{G})
-
\lambda_C C(\mathcal{G}),
\label{eq:score}
\end{equation}
where $\lambda_C$ controls the trade-off between reasoning accuracy and collaboration efficiency.

Successful candidates are further pruned to remove redundant computation.
Starting from a successful topology $\mathcal{G}$, we iteratively remove one edge or removable node and obtain a smaller topology $\mathcal{G}'$.
A pruning operation is accepted only if it reduces the structural cost while preserving task success:
\begin{equation}
\hat{p}_{succ}(\mathcal{G}')
\ge p_{min},
\qquad
C(\mathcal{G}')
<
C(\mathcal{G}),
\end{equation}
where $p_{min}$ is the minimum acceptable success rate.
The resulting compact topology is denoted as $\mathcal{G}^{*}$.

After scoring and pruning, we obtain a retained candidate set
$\mathcal{S}(\mathcal{Q},Y)$ for each training instance.
We then construct a soft supervision distribution over retained candidates:
\begin{equation}
\begin{split}
\tilde{Q}(\mathcal{G}) &= \frac{\exp(\beta_q \mathrm{Score}(\mathcal{G}))}{\sum_{\mathcal{G}'\in\mathcal{S}(\mathcal{Q},Y)} \exp(\beta_q \mathrm{Score}(\mathcal{G}'))}, \
\\&\quad  \mathcal{G}\in\mathcal{S}(\mathcal{Q},Y),
\end{split}
\end{equation}
where $\beta_q$ is a temperature parameter controlling how sharply the distribution concentrates on top-scoring candidates.

Finally, the topology generator is optimized to match this mined supervision distribution:
\begin{equation}
\begin{split}
\mathcal{L}(\theta)
=
\mathbb{E}_{(\mathcal{Q},Y)}
\Bigg[
\sum_{\mathcal{G}\in\mathcal{S}(\mathcal{Q},Y)}
\tilde{Q}(\mathcal{G})\\
\quad \times \Big(
-\log P_\theta(\mathcal{G}\mid \mathcal{Q},\mathcal{C}_{\mathcal{K}})
\Big)
\Bigg]. 
\end{split}
\end{equation}
\vspace{-1em}

Training alternates between candidate mining and generator optimization,
allowing K-GAT to progressively learn evidence-conditioned collaboration structures from execution-derived supervision.

\begin{figure}[t]
    \centering
    \includegraphics[width=1.0\columnwidth]{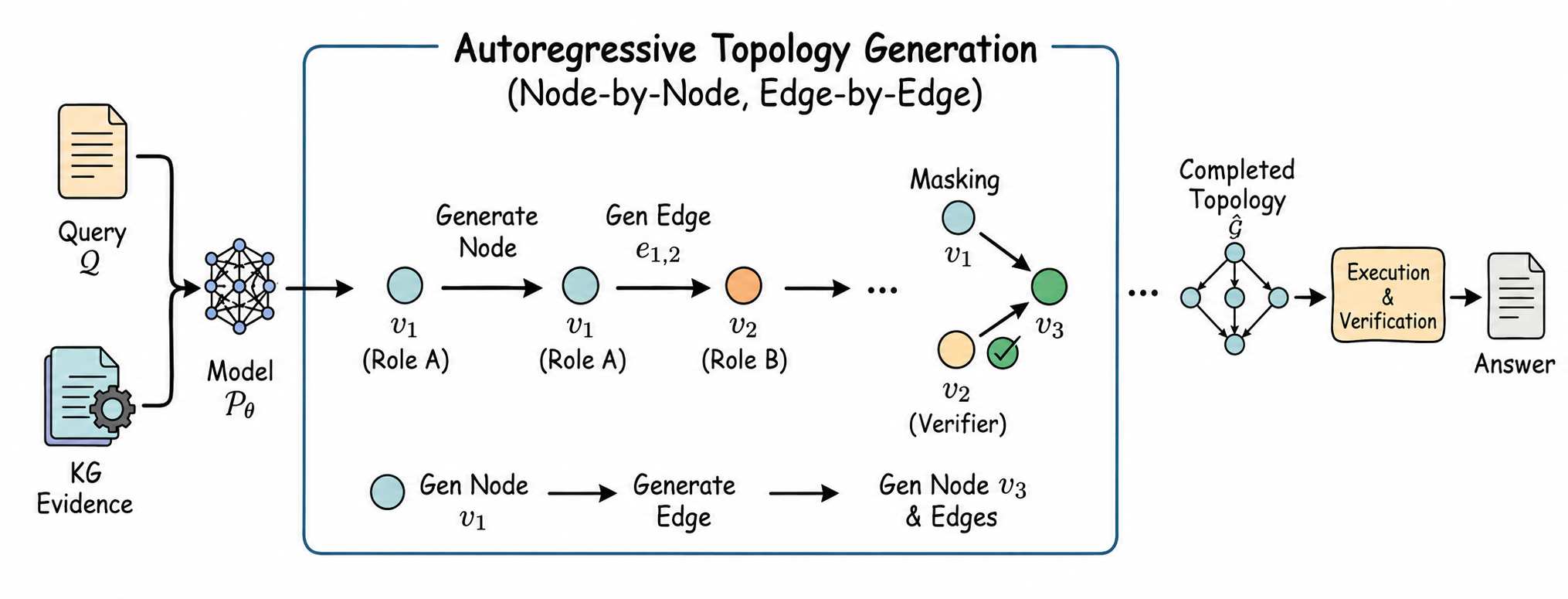}
    \caption{\textbf{The End-to-End Inference Workflow of K-GAT.} The framework initiates by encoding the query $\mathcal{Q}$ and retrieving external evidence. This context is then fed into the model $\mathcal{P}_\theta$ for autoregressive topology generation, which proceeds node-by-node and edge-by-edge under structural masking. The resulting structure $\hat{\mathcal{G}}$ is passed to the execution and synthesize the final answer.}
    \label{fig:inference_pipeline}
\end{figure}
% \vspace{-1em}

\subsection{Dynamic Inference and Execution}
\label{sec:inference}

At test time, K-GAT follows an evidence-first inference pipeline that dynamically adapts the collaboration workflow to the retrieved external knowledge.
Given a test query $\mathcal{Q}_{test}$, the system first retrieves the top-$K$ evidence units and constructs the evidence context
$\mathcal{C}_{\mathcal{K}}^{test}=(\mathcal{T}_{sub},\Pi)$,
where $\mathcal{T}_{sub}$ denotes the retrieved evidence units and $\Pi$ denotes their provenance information.
The retrieved evidence is then linearized together with the query and encoded into an evidence-conditioned representation:
\begin{equation*}
\mathbf{c}_{\mathcal{K}}
=
\mathrm{Pool}\Big(
\mathcal{E}_{ctx}(
\Linearize(\mathcal{T}_{sub},\Pi)\oplus \mathcal{Q}_{test}
)
\Big).
\end{equation*}

Conditioned on $\mathbf{c}_{\mathcal{K}}$,
the topology generator autoregressively constructs a collaboration graph
$\mathcal{G}=(\mathcal{V},\mathcal{E})$.
At generation step $t$, the generator selects the next agent role $v_t$ from the task-specific role inventory $\mathcal{R}_{\mathcal{T}}$ according to the compatibility between the current reasoning state and candidate role representations:
\begin{equation}
\alpha_k
=
\frac{
\exp(
\mathrm{sim}(
\mathbf{r}_k,
[\mathbf{h}_{<t};\mathbf{c}_{\mathcal{K}}]
)
)
}{
\sum_{r_j\in\mathcal{R}_{\mathcal{T}}}
\exp(
\mathrm{sim}(
\mathbf{r}_j,
[\mathbf{h}_{<t};\mathbf{c}_{\mathcal{K}}]
)
)
},
\end{equation}
where $\mathbf{r}_k$ denotes the representation of role $r_k$
and $\mathbf{h}_{<t}$ denotes the representation of the partially generated topology.
The next role is selected according to:
\begin{equation}
v_t=\arg\max_{r_k\in\mathcal{R}_{\mathcal{T}}}\alpha_k.
\end{equation}

After selecting node $v_t$,
K-GAT predicts communication dependencies between $v_t$ and previously generated nodes:
\begin{equation}
P(e_{j,t}=1)
=
\mathrm{MLP}
(
[\mathbf{h}_j;\mathbf{h}_t;\mathbf{c}_{\mathcal{K}}]
),
\end{equation}
where $\mathbf{h}_j$ and $\mathbf{h}_t$ denote the representations of nodes $v_j$ and $v_t$.
During decoding, K-GAT enforces structural constraints including acyclicity, bounded indegree, and valid answer nodes to guarantee executable collaboration workflows.
The decoding process terminates when an END token is generated or the maximum node budget $N_{\max}$ is reached.
The resulting topology is denoted as $\hat{\mathcal{G}}$.

The generated workflow is then executed in topological order.
For each node $v_t$ with role $r_t$, the runtime prompt is constructed from the role-specific instruction, the user query, messages from predecessor agents, and the retrieved evidence:
\begin{align}
m_t = \mathrm{LLM}\Big(
&\mathcal{S}^{(r_t)}_{sys}
\oplus \mathcal{Q}_{test}
\oplus \mathrm{Agg}(\{m_j\}_{j\in\mathcal{N}_{in}(v_t)}) \notag\\
&\oplus \Linearize(\mathcal{T}_{sub},\Pi)
\Big),
\end{align}
where $\mathcal{N}_{in}(v_t)$ denotes the predecessor nodes of $v_t$ and
$\mathrm{Agg}(\cdot)$ aggregates their intermediate messages.

\subsection{Knowledge-Grounded Verification}

To reduce error propagation during multi-agent collaboration,
K-GAT incorporates a KG-Verifier module during execution.
The verifier checks intermediate agent responses against provenance-linked evidence retrieved from the external knowledge source.
When unsupported or inconsistent statements are detected,
the verifier either revises the intermediate response or terminates the current topology execution.
The verification process additionally records provenance-linked evidence spans and correction traces.

\subsection{Complexity Analysis}

The inference cost of K-GAT consists of three components:
evidence retrieval, topology decoding, and multi-agent execution.
After indexing, retrieval scales linearly with the number of retrieved evidence units $K_{ret}$.
Given a generated topology with at most $N_{\max}$ nodes and bounded indegree $d_{\max}$,
topology decoding requires $O(N\cdot d_{\max})$ edge predictions.
The overall execution cost further scales linearly with the number of instantiated agents and their prompt lengths.
Detailed complexity analysis is provided in Appendix~\ref{app:complexity}.

\begin{table*}[t]
    \centering
    \caption{\textbf{Main Results across Seven Reasoning Benchmarks.} We report accuracy (\%) for all methods. Comparison methods in the bottom three sections utilize the same \textsc{Qwen-3-8B} backbone to ensure fair comparison. The best performance is formatted in \textbf{bold}, while the second-best is underlined.}
    \label{tab:main_results_full}
    \resizebox{\textwidth}{!}{
    \begin{tabular}{lcccc|ccc|c}
        \toprule
        & \multicolumn{4}{c|}{\textbf{Knowledge-Intensive Reasoning}} & \multicolumn{3}{c|}{\textbf{Symbolic \& Algorithmic}} & \\
        \cmidrule(lr){2-5} \cmidrule(lr){6-8}
        \textbf{Method} & \textbf{MMLU} & \textbf{MMLU-Pro} & \textbf{GPQA} & \textbf{StratQA} & \textbf{GSM8K} & \textbf{AQuA} & \textbf{HumanEval} & \textbf{Avg.} \\
        \midrule
        \multicolumn{9}{l}{\textit{\textbf{Large-Scale Single Models}}} \\
        \textsc{Mistral-8$\times$7B} & 70.60 & 20.36 & 22.32 & 68.34 & 69.45 & 58.30 & 40.20 & 49.94 \\
        \textsc{Qwen-3-32B} & 83.61 & 72.41 & 49.26 & 77.60 & 78.09 & 85.43 & 78.50 & 74.99 \\
        \textsc{Llama-3.1-70B} & 83.60 & 66.40 & 46.70 & 75.28 & \textbf{95.10} & 67.32 & 80.50 & 73.56 \\
        \midrule
        \multicolumn{9}{l}{\textit{\textbf{Static Collaboration Topologies (Backbone: Qwen-3-8B)}}} \\
        \textsc{Chain} & 80.70 & 41.96 & 34.96 & 67.34 & 85.67 & \textbf{87.40} & 71.34 & 67.05 \\
        \textsc{Star} & 78.94 & 44.28 & 35.26 & 72.05 & 78.77 & 85.03 & 73.78 & 66.87 \\
        \textsc{Tree} & 78.59 & 46.61 & 37.50 & 57.20 & 86.56 & 82.59 & 73.17 & 66.03 \\
        \midrule
        \multicolumn{9}{l}{\textit{\textbf{Multi-Agent System Frameworks (Backbone: Qwen-3-8B)}}} \\
        \textsc{LLM-Debate} & 80.70 & 47.32 & 35.04 & 65.37 & 86.80 & 78.74 & 79.26 & 67.60 \\
        \textsc{AgentPrune} & 79.29 & 47.67 & 35.71 & 65.45 & 88.70 & 82.67 & 79.87 & 68.48 \\
        \textsc{AgentDropout} & 77.19 & 46.78 & 37.27 & 65.67 & 90.92 & 79.52 & 81.09 & 68.35 \\
        \textsc{AFlow} & 81.70 & 47.85 & 36.11 & 67.33 & 90.31 & 81.89 & 81.98 & 69.60 \\
        \textsc{G-Designer} & 81.05 & 48.21 & 36.83 & 67.68 & 90.97 & 83.07 & 82.31 & 70.02 \\
        \midrule
        \multicolumn{9}{l}{\textit{\textbf{Ours (Backbone: Qwen-3-8B)}}} \\
        \textsc{K-GAT} (Inf w/o KG) & \underline{83.66} & \underline{52.50} & \underline{40.19} & \underline{68.25} & 91.95 & \underline{85.82} & \underline{84.29} & \underline{72.38} \\
        \textsc{K-GAT} (Ours) & \textbf{87.71} & \textbf{66.42} & \textbf{50.75} & \textbf{84.97} & \underline{91.96} & 84.23 & \textbf{84.75} & \textbf{78.68} \\
        \bottomrule
    \end{tabular}
    }
\end{table*}

\section{Experiment}

\subsection{Experiment Setups}

\paragraph{Datasets}

We evaluate K-GAT across seven benchmarks categorized into two primary modalities to assess both evidence-grounded knowledge application and precise symbolic logic. \textbf{Knowledge-Intensive Reasoning.} A core objective of K-GAT is to handle queries requiring robust external grounding. To this end, we employ four knowledge-centric datasets: \textsc{MMLU}~\cite{hendrycks2021ethics}, \textsc{MMLU-Pro}~\cite{wang2024mmlu}, \textsc{GPQA}~\cite{rein2024gpqa} and \textsc{StrategyQA}~\cite{geva2021did}. \textbf{Symbolic and Algorithmic Reasoning.} Complementing knowledge retrieval, we assess the framework's capacity for structural deduction and syntactic precision. We utilize \textsc{GSM8K}~\cite{cobbe2021gsm8k} and \textsc{AQuA}~\cite{ling2017program} to evaluate multi-step arithmetic and algebraic derivation capabilities. Furthermore, \textsc{HumanEval}~\cite{chen2021evaluating} is included to test algorithmic logic and code generation proficiency.

\paragraph{Baselines}

We evaluate K-GAT against three distinct categories of baselines. \textbf{Large-Scale Single Models.} We benchmark against representative open-weight LLMs, including \textsc{Mistral-8$\times$7B}, \textsc{Qwen-3-32B}~\cite{yang2025qwen3}, and \textsc{Llama-3.1-70B}~\cite{dubey2024llama}. \textbf{Static Collaboration Topologies.} We evaluate performance against fundamental fixed structures, specifically \textsc{Chain}, \textsc{Star}, and \textsc{Tree}. \textbf{Multi-Agent System Frameworks.} We further compare against established MAS paradigms, including \textsc{LLM-Debate}~\cite{liang2023encouraging}, \textsc{AgentPrune}~\cite{zhang2024cut}, \textsc{AgentDropout}~\cite{wang2025agentdropout}, and the generative baseline \textsc{AFlow}~\cite{zhang2025aflow}, \textsc{G-Designer}~\cite{zhang2024g}.

\paragraph{Implementation Details}

We primarily utilize \textsc{Qwen-3-8B} as the backbone for both the autoregressive topology generator and the instantiated agents. The external knowledge source is constructed from large-scale Wikipedia corpora using the StructSense pipeline~\cite{chhetri2025structsense}, detailed construction procedures are provided in Appendix~\ref{app:kg_construction}. To support multidisciplinary reasoning, we provide explicit agent profiles for different tasks to guide their collaborative behavior; the detailed prompts for these roles are provided in Appendix~\ref{sec:appendix_profiles}. For evidence retrieval, the context encoder is initialized with \textsc{all-MiniLM-L6-v2}, with the embedding dimension set to $D=384$. For each query, we retrieve the top-$3$ relevant documents as external evidence.
Regarding structural constraints, we set the maximum topology size to $N_{\max}=6$ and the indegree bound to $d_{\max}=3$ to balance structural expressiveness and computational efficiency. For curriculum optimization, we train the generator using the AdamW optimizer with a learning rate of $2\mathrm{e}{-5}$ and a batch size of 4, employing a cosine annealing schedule over the training iterations.

\subsection{Overall Performance}
\label{sec:analysis}

The results in Table~\ref{tab:main_results_full} demonstrate that K-GAT consistently achieves the strongest overall performance among all 8B-scale baselines, attaining an average accuracy of 78.68\%. Notably, despite using only an 8B backbone, K-GAT remains competitive with substantially larger models such as \textsc{Qwen-3-32B} and \textsc{Llama-3.1-70B}. Moreover, \textsc{K-GAT (Inf w/o KG)} still achieves strong performance even without external knowledge retrieval during inference, obtaining an average accuracy of 72.38\% and consistently outperforming existing MAS baselines. This suggests that the gains of K-GAT are not solely dependent on retrieved knowledge. Instead, incorporating knowledge-grounded workflows during training enables the model to internalize more effective collaboration behaviors, such as verification and adaptive coordination, which remain beneficial even in closed-book inference settings.

We further observe that the performance gains are particularly significant on knowledge-intensive reasoning tasks. In contrast, the improvements on symbolic or algorithmic tasks such as \textsc{GSM8K} are comparatively smaller, since these tasks rely less on external knowledge retrieval and more on intrinsic mathematical or procedural reasoning abilities of the backbone model.
Additional experiments with a stronger \textsc{Qwen3-235B-A22B} backbone are provided in Appendix~\ref{app:stronger_backbone}.

\subsection{Ablation and Component Analysis}
\label{sec:ablation}

% --- Paragraph 1: Focus on Internal Mechanisms (Table 1) ---
To strictly validate the proposed framework, we first conduct a comprehensive ablation analysis focusing on component efficacy (Table~\ref{tab:ablation_kg_components}). Regarding internal mechanisms, we observe that the topology generator trained with our curriculum achieves a performance gain even in the absence of external evidence (40.19\% vs. 38.48\%), suggesting the successful internalization of reasoning structures; however, the peak performance of \textbf{50.75\%} is only attained when combining the evidence-aware topology with the \textsc{KG-Verifier}, which effectively filters hallucinated paths.

% --- Table 1 Code ---
\begin{table}[h!] % 使用 h! 尝试让表格紧跟在第一段文字后
    \centering
    \caption{\textbf{Ablation Study on Knowledge Integration and Verification.} We analyze the impact of incorporating Knowledge Graphs (KG) during the training and inference phases, as well as the specific contribution of the KG-Verifier on the GPQA dataset.}
    \label{tab:ablation_kg_components}
    \resizebox{\columnwidth}{!}{
    \begin{tabular}{ccc|c}
        \toprule
        \textbf{Training} & \multicolumn{2}{c|}{\textbf{Inference Phase}} & \textbf{Performance} \\
        \cmidrule(lr){2-3}
        \textbf{w/ KG} & \textbf{w/ KG} & \textbf{w/ Verifier} & \textbf{GPQA (\%)} \\
        \midrule
        $-$ & $-$ & $-$ & 38.48 \\
        $-$ & \checkmark & $-$ & 49.26 \\
        $-$ & \checkmark & \checkmark & 50.00 \\
        \midrule
        \checkmark & $-$ & $-$ & 40.19 \\
        \checkmark & \checkmark & $-$ & 50.00 \\
        \checkmark & \checkmark & \checkmark & \textbf{50.75} \\
        \bottomrule
    \end{tabular}
    }
\end{table}

% --- Paragraph 2: Focus on Leakage Analysis (Table 2) ---
Complementing the component ablation, we further examine whether the performance gains of K-GAT genuinely arise from utilizing external knowledge rather than relying on memorized parametric information. Table~\ref{tab:leakage_analysis} compares closed-book and open-book performance on MMLU-Pro and GPQA. Across both datasets, K-GAT exhibits the largest improvement after incorporating external knowledge, achieving gains of +15.35\% and +10.56\%, respectively. These gains are consistently larger than those of single-agent baselines and the query-conditioned G-Designer, suggesting that K-GAT more effectively converts retrieved evidence into useful collaborative reasoning signals.

\begin{table*}[t]
    \centering
    \caption{\textbf{Knowledge Dependence Analysis.} We compare closed-book and open-book performance across single-agent and dynamic multi-agent settings. $\Delta$ denotes the performance gain obtained after incorporating external knowledge.}
    \label{tab:leakage_analysis}
    \resizebox{\textwidth}{!}{
    \begin{tabular}{ll|l|cc|c}
        \toprule
        \textbf{Dataset} & \textbf{Method} & \textbf{Agent/Topology Setting} 
        & \textbf{w/o KG} & \textbf{w/ KG} & \textbf{Gain ($\Delta$)} \\
        \midrule

        \multirow{4}{*}{\textsc{MMLU-Pro}}
        & \textsc{Mistral-8$\times$7B} 
        & Direct reasoning, single agent
        & 20.36 & 25.00 & +4.64\% \\

        & Qwen-3-8B (Base)
        & Direct reasoning, single agent
        & 38.50 & 46.20 & +7.70\% \\

        & \textsc{G-Designer}
        & Query-conditioned dynamic multi-agent topology
        & 48.21 & 58.92 & +10.71\% \\

        & \textbf{K-GAT (Ours, Qwen-3-8B)}
        & \textbf{Evidence-conditioned dynamic multi-agent topology}
        & \textbf{51.07} & \textbf{66.42} & \textbf{+15.35\%} \\
        \midrule

        \multirow{4}{*}{\textsc{GPQA}}
        & \textsc{Mistral-8$\times$7B}
        & Direct reasoning, single agent
        & 22.32 & 26.96 & +4.64\% \\

        & Qwen-3-8B (Base)
        & Direct reasoning, single agent
        & 33.09 & 39.22 & +6.13\% \\

        & \textsc{G-Designer}
        & Query-conditioned dynamic multi-agent topology
        & 36.83 & 45.34 & +8.51\% \\

        & \textbf{K-GAT (Ours, Qwen-3-8B)}
        & \textbf{Evidence-conditioned dynamic multi-agent topology}
        & \textbf{40.19} & \textbf{50.75} & \textbf{+10.56\%} \\

        \bottomrule
    \end{tabular}
    }
\end{table*}

To further disentangle the contributions of retrieval, topology selection, and verification, we conduct a controlled comparison on \textsc{GPQA}, as shown in Table~\ref{tab:controlled_component_analysis}. All open-book variants use Qwen-3-8B as the common backbone and receive the same top-3 retrieved evidence. The KG-Verifier is disabled when comparing topology strategies and is introduced only in the final configuration, allowing its contribution to be evaluated separately.

\begin{table*}[t]
    \centering
    \caption{\textbf{Controlled Component Analysis on GPQA.} All open-book settings use Qwen-3-8B and identical top-3 retrieved evidence. The verifier is disabled when comparing topology strategies and enabled only in the full K-GAT setting.}
    \label{tab:controlled_component_analysis}
    \renewcommand{\arraystretch}{1.1} % 稍微拉开行高
    \begin{tabular*}{\textwidth}{@{\extracolsep{\fill}} lccc|c }
        \toprule
        \textbf{Setting} & \textbf{External KG} & \textbf{Topology} & \textbf{KG-Verifier} & \textbf{GPQA (\%)} \\
        \midrule
        Qwen-3-8B (Base)
        & $-$ & None & $-$ & 33.09 \\

        Qwen-3-8B + KG
        & \checkmark & None & $-$ & 39.22 \\

        \textsc{G-Designer} + KG
        & \checkmark & Query-conditioned & $-$ & 45.34 \\

        K-GAT w/o Verifier
        & \checkmark & Evidence-conditioned & $-$ & 50.00 \\

        \textbf{K-GAT (Full)}
        & \checkmark & \textbf{Evidence-conditioned} & \checkmark & \textbf{50.75} \\
        \bottomrule
    \end{tabular*}
\end{table*}

The comparison from 33.09\% to 39.22\% quantifies the gain from retrieval alone. With the retrieved evidence and verifier setting fixed, introducing a query-conditioned topology further improves accuracy to 45.34\%, while the evidence-conditioned topology of K-GAT reaches 50.00\%. The resulting 4.66-point improvement over \textsc{G-Designer} isolates the benefit of conditioning collaboration structure on the retrieved evidence rather than solely on the query. Enabling the KG-Verifier further increases performance from 50.00\% to 50.75\%, indicating an additional but distinct contribution from knowledge-grounded verification. Thus, the controlled setting separates the effects of retrieval, topology generation, and verification while keeping the backbone and external evidence unchanged. Curriculum optimization and pruning are training mechanisms for learning the topology generator and are not individually isolated in this comparison.

We also note that the maximum topology size $N_{\max}=6$ is a shared constraint across all compared methods.
To examine whether this node budget limits the learned collaboration structures, we conduct a sensitivity analysis on \textsc{StrategyQA}, as shown in Table~\ref{tab:nmax_analysis}.
The results indicate that performance is not constrained by the topology size limit: increasing $N_{\max}$ does not improve accuracy, while the average generated topology consistently remains around 2.34 nodes.
This suggests that a small node budget is already sufficient for K-GAT to learn effective collaboration topologies in this setting.

\begin{table}[h]
\centering
\small
\caption{Sensitivity analysis of topology size limit $N_{\max}$ on \textsc{StrategyQA}.}
\label{tab:nmax_analysis}
\begin{tabular}{ccc}
\toprule
$N_{\max}$ & Accuracy & Avg. Nodes \\
\midrule
6 & 84.97 & 2.34 \\
7 & 84.95 & 2.35 \\
8 & 84.97 & 2.34 \\
\bottomrule
\end{tabular}
\end{table}

\subsection{Evidence-Conditioned Topology Adaptation}
\label{sec:topology_adaptation}

To examine whether K-GAT adapts its collaboration structure to different evidence conditions, we manually analyze 100 \textsc{StrategyQA} instances and group them according to the quality of the retrieved evidence. As shown in Table~\ref{tab:evidence_topology_adaptation}, highly relevant and consistent evidence typically leads to compact two-node workflows, whereas conflicting or low-relevance evidence results in larger and more diverse topologies. The average topology size increases from 2.02 to 2.66 nodes under challenging evidence conditions, indicating that K-GAT dynamically allocates additional collaboration and verification when the retrieved evidence is less reliable.

\begin{table}[t]
    \centering
    \caption{\textbf{Topology Adaptation under Different Evidence Conditions.} Distribution of generated topology sizes over 100 manually analyzed \textsc{StrategyQA} instances.}
    \label{tab:evidence_topology_adaptation}
    \resizebox{\columnwidth}{!}{
    \begin{tabular}{l|c|ccc|c}
        \toprule
        \textbf{Evidence Condition} & \textbf{Cases} &
        \textbf{2 Nodes} & \textbf{3 Nodes} & \textbf{4 Nodes} &
        \textbf{Avg. Nodes} \\
        \midrule
        Highly relevant and consistent
        & 50 & 49 & 1 & 0 & 2.02 \\
        Conflicting or low-relevance
        & 50 & 23 & 21 & 6 & 2.66 \\
        \midrule
        Overall
        & 100 & 72 & 22 & 6 & 2.34 \\
        \bottomrule
    \end{tabular}
    }
\end{table}

\subsection{Computational Efficiency}
\label{sec:efficiency}

\begin{figure}[htbp]
    \centering
    \begin{subfigure}{0.48\linewidth}
        \centering
        \includegraphics[width=\linewidth]{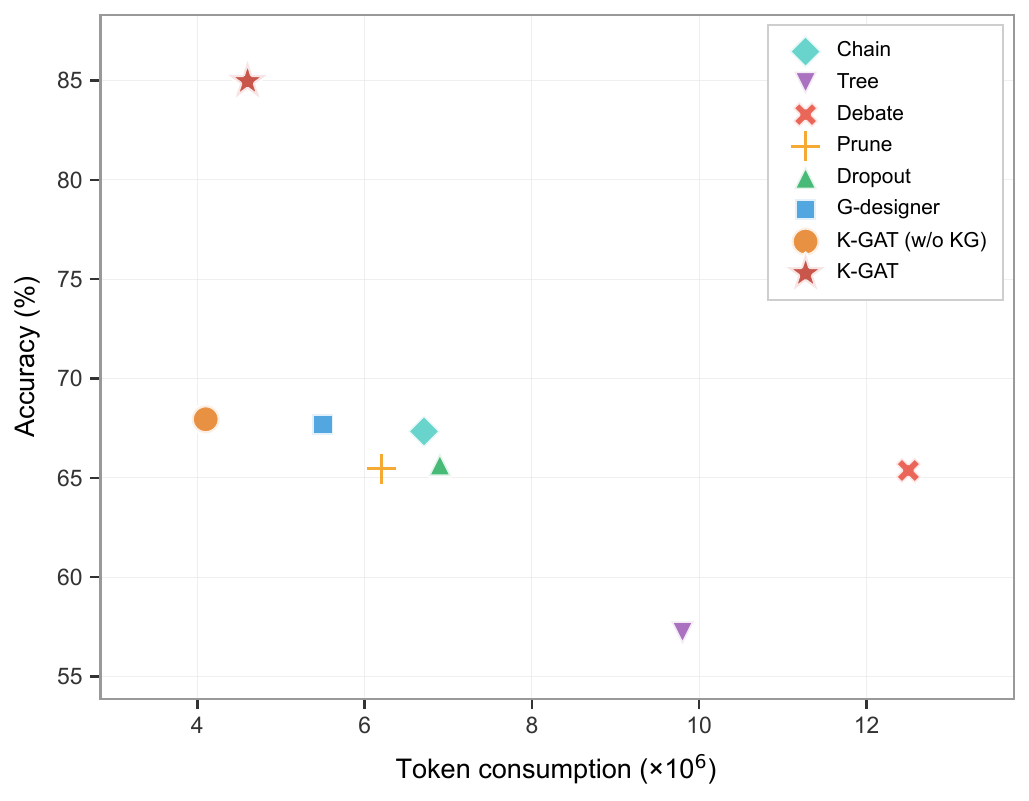}
        \caption{StrategyQA}
        \label{fig:eff_strategyqa}
    \end{subfigure}
    \hfill
    \begin{subfigure}{0.48\linewidth}
        \centering
        \includegraphics[width=\linewidth]{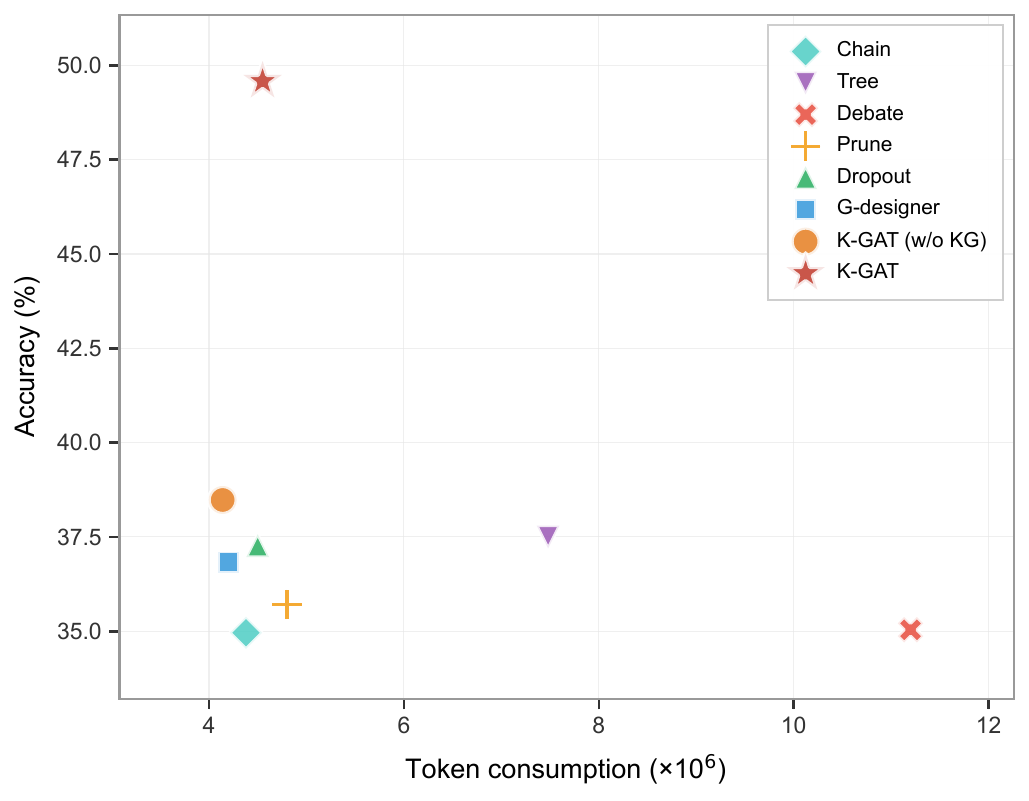}
        \caption{GPQA}
        \label{fig:eff_gpqa}
    \end{subfigure}
    \caption{\textbf{Cost-Performance Analysis.} The plot compares accuracy ($y$-axis) against total token consumption ($x$-axis). K-GAT (Red Star) achieves a superior trade-off, significantly reducing computational costs compared to \textsc{LLM-Debate} and \textsc{Tree} while maintaining high performance.}
    \label{fig:efficiency}
\end{figure}

We evaluate the cost-effectiveness of K-GAT by visualizing the trade-off between accuracy and total token consumption in Figure~\ref{fig:efficiency}. As illustrated, K-GAT consistently resides in the optimal top-left region across both datasets, indicating high accuracy with low computational cost. Specifically, compared to the computationally intensive \textsc{LLM-Debate} on GPQA, our method reduces token consumption by over 50\% while achieving higher accuracy. Similarly, on StrategyQA, K-GAT avoids the extensive search overhead of the \textsc{Tree} topology, maintaining an efficiency profile comparable to simple linear chains but with significantly superior reasoning performance.

\begin{figure}[htbp]
    \centering
    \includegraphics[width=\linewidth]{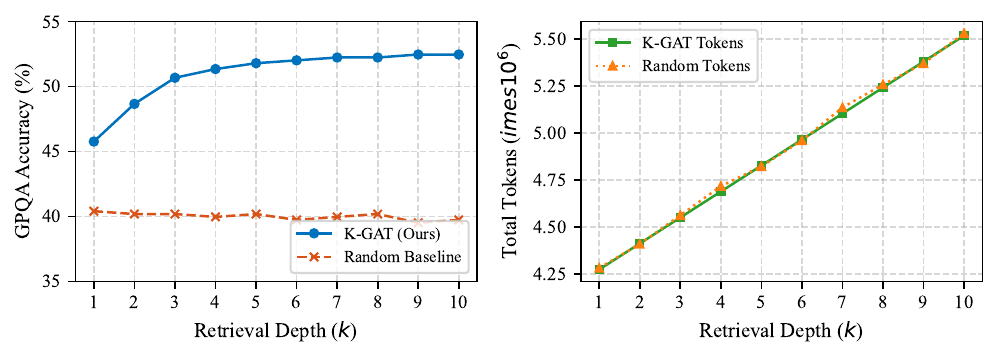}
    \vspace{-1em}
    \caption{\textbf{Sensitivity Analysis on Retrieval Depth ($K$).} The left panel illustrates the GPQA accuracy trend as $K$ increases from 1 to 10, comparing K-GAT against a random retrieval baseline. The right panel depicts the corresponding total token consumption.}
    \vspace{-1em}
    \label{fig:sensitivity}
\end{figure}

\vspace{-1em}
\subsection{Comparison with ReAct-Style Reasoning}
\label{sec:react_analysis}

To better understand the advantages of evidence-conditioned topology generation, we compare K-GAT with a ReAct-style reasoning framework under the same external knowledge source. As shown in Table~\ref{tab:react_comparison}, K-GAT consistently outperforms ReAct-style reasoning. Unlike ReAct, which determines actions sequentially through local step-wise decisions, K-GAT explicitly plans the global collaboration structure based on the retrieved evidence before execution. This enables adaptive allocation of verification and collaboration behaviors according to the evidence condition, leading to more reliable reasoning on knowledge-intensive tasks.

\begin{table}[h]
\centering
\small
\caption{Comparison between ReAct-style reasoning and K-GAT.}
\label{tab:react_comparison}
\resizebox{\columnwidth}{!}{
\begin{tabular}{lcccc}
\toprule
Method & Style & GPQA & MMLU-Pro & StrategyQA \\
\midrule
ReAct-style & Step-wise & 31.25 & 56.25 & 68.52 \\
K-GAT & Evidence-conditioned & \textbf{50.75} & \textbf{66.42} & \textbf{84.97} \\
\bottomrule
\end{tabular}
}
\end{table}

\vspace{-1em}

\subsection{Sensitivity Analysis: Retrieval Depth}
\label{sec:sensitivity}

We investigate the impact of retrieval depth $K$ on the GPQA dataset in Figure~\ref{fig:sensitivity}. In addition to standard retrieval, we include a random retrieval baseline to simulate low-quality or mismatched knowledge graph conditions. K-GAT demonstrates a clear accuracy improvement as $K$ increases, while the random retrieval baseline remains largely stagnant, indicating that the gains mainly originate from relevant evidence rather than simply increasing the context length. However, since token consumption scales linearly with retrieval depth and performance saturates beyond $K=6$, we identify $K \in [3,5]$ as the best trade-off between reasoning performance and computational cost.

\section{Conclusion}
We identify \textit{Structural Mismatch} in dynamic multi-agent systems, where workflows are generated solely from query semantics before observing evidence. To address this, we propose \textbf{K-GAT}, an evidence-first framework that generates collaboration topologies conditioned on retrieved evidence.

\section*{Limitations}

K-GAT relies on an external knowledge graph to support evidence-conditioned topology generation. Therefore, the effectiveness of the framework depends on the quality and coverage of the underlying knowledge source. In this work, we construct a large-scale provenance-grounded knowledge graph from Wikipedia using the StructSense pipeline, which introduces additional preprocessing and storage costs compared with purely parametric reasoning systems. Moreover, when the retrieved evidence is incomplete or noisy, the generated collaboration topology may still become suboptimal.

In addition, the current framework mainly focuses on relatively compact collaboration structures with bounded topology size, which may limit scalability for more complex long-horizon reasoning scenarios. Finally, although K-GAT demonstrates strong performance on several knowledge-intensive benchmarks, extending evidence-conditioned topology generation to broader tool-use environments and dynamically evolving knowledge sources remains an important direction for future work.

\section*{Ethical Considerations}

This work uses publicly available datasets, open-source language models, and external knowledge sources constructed from Wikipedia corpora. All utilized artifacts are restricted to academic research purposes under their respective open-source licenses. We do not involve private user data or human annotations.

\section*{Acknowledgements}

This research was supported by \textit{Hubei Provincial Department of Science and Technology} (No. 2025AFB144), \textit{National Natural Science Foundation of China} (No. 62406121), and \textit{Natural Science Foundation of Hubei Province, China} (No. 2024AFB189).

\bibliography{custom}

% \newpage

\appendix

\section{Notation Summary}
\label{app:notation}

\begin{table*}[t]
\centering
\small
\caption{Summary of major symbols used in K-GAT.}
\label{tab:notation}
\begin{tabular}{p{2.2cm} p{11.8cm}}
\toprule
\textbf{Symbol} & \textbf{Description} \\
\midrule

$\mathcal{Q}$ & Input query. \\

$Y$ & Ground-truth answer corresponding to query $\mathcal{Q}$. \\

$\mathcal{D}$ & Training dataset consisting of query--answer pairs. \\

$\mathcal{R}_{\mathcal{T}}$ & Task-specific agent role inventory. \\

$r_t$ & Agent role assigned to node $v_t$. \\

$\mathcal{S}_{sys}^{(r)}$ & System prompt associated with role $r$. \\

$\mathcal{T}_{sub}$ & Retrieved evidence units from the external knowledge source. \\

$\Pi$ & Provenance information associated with retrieved evidence. \\

$\mathcal{C}_{\mathcal{K}}$ & Evidence context formed by retrieved evidence and provenance information. \\

$\mathbf{q}$ & Embedding representation of the query. \\

$\mathbf{x}_{\tau}$ & Embedding representation of evidence unit $\tau$. \\

$s(\tau;\mathcal{Q})$ & Similarity score between query $\mathcal{Q}$ and evidence unit $\tau$. \\

$\mathcal{G}=(\mathcal{V},\mathcal{E})$ & Collaboration topology represented as a directed acyclic graph (DAG). \\

$\mathcal{V}$ & Set of instantiated agent nodes. \\

$\mathcal{E}$ & Set of communication edges between agents. \\

$v_t$ & Agent node generated at step $t$. \\

$e_{j,t}$ & Directed communication edge between nodes $v_j$ and $v_t$. \\

$\mathbf{h}_{<t}$ & Representation of the partially generated topology before step $t$. \\

$\mathbf{c}_{\mathcal{K}}$ & Encoded evidence-conditioned representation. \\

$P_\theta(\mathcal{G}\mid\mathcal{Q},\mathcal{C}_{\mathcal{K}})$ & Conditional topology generation distribution parameterized by $\theta$. \\

$C(\mathcal{G})$ & Structural cost of topology $\mathcal{G}$. \\

$\lambda_V,\lambda_E$ & Cost coefficients for nodes and edges. \\

$\hat{p}_{succ}(\mathcal{G})$ & Empirical execution success rate of topology $\mathcal{G}$. \\

$\mathrm{Score}(\mathcal{G})$ & Final candidate score combining execution success and structural efficiency. \\

$\mathcal{S}(\mathcal{Q},Y)$ & Retained candidate topology set for training instance $(\mathcal{Q},Y)$. \\

$\tilde{Q}(\mathcal{G})$ & Soft supervision distribution over retained candidate topologies. \\

$\beta_q$ & Temperature parameter controlling supervision sharpness. \\

$\mathcal{L}(\theta)$ & Distribution-matching training objective for topology generation. \\

$N_{\max}$ & Maximum allowed topology size during decoding. \\

$d_{\max}$ & Maximum indegree constraint for each node. \\

\bottomrule
\end{tabular}
\end{table*}

\section{Complexity Analysis and Token Accounting}
\label{app:complexity}

This section provides a complexity analysis of K-GAT, including evidence retrieval, topology decoding, verifier operations, and end-to-end token consumption during inference.

\subsection{Notation}

Let $K_{ret}$ denote the number of retrieved evidence units, and let
$N=|\mathcal{V}|\le N_{\max}$ denote the number of generated agent nodes.
We further denote by $d_{\max}$ the maximum indegree constraint,
$R$ the repeated execution count used during training-time candidate evaluation,
$L$ the number of extracted atomic claims used by the verifier,
and $K_{nbr}$ the number of retrieved neighboring evidence units for each claim.
We assume the evidence embeddings are indexed using an approximate nearest neighbor (ANN) structure such as HNSW.

\subsection{Retrieval Complexity}

Given a query embedding, retrieving the top-$K_{ret}$ evidence units from the ANN index requires:
\begin{equation}
T_{\text{ret}}
=
T_{ann}(K_{ret})
+
O(K_{ret}),
\end{equation}
where $T_{ann}(K_{ret})$ denotes the ANN retrieval cost and the linear term corresponds to evidence collection and provenance assembly.

After retrieval, provenance-linked evidence spans are attached to the retrieved evidence units.
Assuming an average of $\bar{P}$ provenance spans per evidence unit and entailment scoring cost $T_{\mathrm{Entail}}$, the provenance verification cost is:
\begin{equation}
T_{\text{prov}}
=
O(K_{ret}\cdot \bar{P}\cdot T_{\mathrm{Entail}}).
\end{equation}

In practice, provenance spans are precomputed offline, making the runtime overhead relatively small.

\subsection{Topology Decoding Complexity}

K-GAT autoregressively generates the collaboration topology by predicting both agent nodes and communication edges.

At each generation step, the next node is selected from the role inventory $\mathcal{R}_{\mathcal{T}}$, leading to:
\begin{equation}
T_{\text{node}}
=
O(N\cdot |\mathcal{R}_{\mathcal{T}}|).
\end{equation}

For each generated node, K-GAT additionally predicts communication edges to previously generated nodes under the indegree constraint $d_{\max}$:
\begin{equation}
T_{\text{edge}}
=
O\!\left(
N\cdot \min\{N,\alpha_d d_{\max}\}
\right),
\end{equation}
where $\alpha_d$ denotes the effective edge sparsity factor introduced by structural masking.

If beam search with beam width $B$ is used during decoding, the overall topology decoding complexity becomes:
\begin{equation}
T_{\text{decode}}
=
O\!\left(
B\cdot
(T_{\text{node}}+T_{\text{edge}})
\right).
\end{equation}

\subsection{Verifier Complexity}

When a generated topology contains a KG-Verifier node, the verifier extracts atomic claims from intermediate responses and checks them against retrieved evidence.

Let $T_{\text{extract}}(m)$ denote the claim extraction cost for message $m$.
The overall verification complexity is:
\begin{equation}
\begin{aligned}
T_{\text{ver}}
=&\ T_{\text{extract}}(m)
+
\sum_{\ell=1}^{L}
\Big(
O(K_{ret}) \\
&\quad+
O(K_{nbr}\cdot \bar{P}\cdot T_{\mathrm{Entail}})
\Big).
\end{aligned}
\end{equation}

Since verification is only triggered for verifier nodes, this overhead scales with the generated topology rather than being applied globally to all reasoning steps.

\subsection{Token Consumption}

For each agent node $v_t$, the runtime prompt consists of the query, retrieved evidence, and aggregated messages from predecessor nodes.
The prompt length is therefore bounded by:
\begin{equation}
L_{\text{prompt}}(t)
\le
L_Q
+
L_K
+
d_{\max}\bar{L}_M,
\end{equation}
where $L_Q$ denotes the query length,
$L_K$ denotes the evidence context length,
and $\bar{L}_M$ denotes the average intermediate message length.

The total token consumption of the generated workflow can then be approximated as:
\begin{equation}
\begin{split}
L_{\text{total}}
\lesssim
N(L_Q+L_K+d_{\max}\bar{L}_M)
\\+
N\bar{L}_{gen}
+
N_{ver}\bar{L}_{ver}, 
\end{split}
\end{equation}
where $\bar{L}_{gen}$ denotes the average generated response length per agent,
$N_{ver}$ denotes the number of verifier nodes,
and $\bar{L}_{ver}$ denotes the average verification output length.

Overall, the inference cost of K-GAT scales approximately linearly with the number of generated agents and retrieved evidence units, while the bounded indegree constraint prevents excessive communication overhead in large collaboration topologies.
% =========================================================
\FloatBarrier

\section{Training Cost of Curriculum Optimization}
\label{sec:training_cost}

Although K-GAT is designed to reduce inference-time collaboration overhead, its curriculum optimization requires repeated execution of candidate topologies to construct execution-based supervision. We therefore provide a detailed breakdown of the training cost in this section.

For each dataset, we construct a fixed curriculum subset of 40 representative instances using stratified sampling, rather than evaluating candidate topologies over the entire training set. For each instance, five initial candidate topologies are generated, and each candidate is executed twice ($R=2$) to estimate its empirical success rate. This results in 400 initial MAS executions per dataset, followed by additional executions introduced by topology pruning and validation.

Table~\ref{tab:curriculum_stage_cost} decomposes the end-to-end wall-clock cost of curriculum optimization. The reported four-to-five-hour training time includes candidate construction, initial candidate execution and scoring, pruning with re-execution, and topology-generator optimization.

\begin{table*}[t]
    \centering
    \caption{\textbf{Stage-wise Training Cost of Curriculum Optimization.}
    We report the wall-clock cost of each stage on \textsc{AQuA} and \textsc{GPQA}.}
    \label{tab:curriculum_stage_cost}
    \resizebox{\textwidth}{!}{
    \begin{tabular}{l|l|cc}
        \toprule
        \textbf{Stage} & \textbf{Main Operation} & \textbf{AQuA} & \textbf{GPQA} \\
        \midrule

        Candidate construction
        & Evidence retrieval and construction of five initial candidate topologies per instance
        & $\approx$ 0.10 h
        & $\approx$ 0.12 h \\

        Initial execution and scoring
        & Two complete MAS executions per candidate ($R=2$) and empirical success-rate estimation
        & $\approx$ 1.60 h
        & $\approx$ 1.95 h \\

        Pruning and re-execution
        & Generation, connectivity checking, and execution-based validation of feasible pruned topologies
        & $\approx$ 2.10 h
        & $\approx$ 2.65 h \\

        Generator optimization
        & Construction of the supervision distribution and distribution-matching optimization
        & $\approx$ 0.20 h
        & $\approx$ 0.28 h \\

        \midrule
        \textbf{Total wall-clock time}
        & \textbf{End-to-end curriculum optimization}
        & \textbf{$\approx$ 4.00 h}
        & \textbf{$\approx$ 5.00 h} \\

        \bottomrule
    \end{tabular}
    }
\end{table*}

Most of the training cost originates from execution-based supervision collection rather than gradient-based optimization. In particular, the pruning stage is slightly more expensive than the initial candidate-evaluation stage because pruning may generate multiple feasible simplified variants. These variants must first satisfy structural and connectivity constraints and are then re-executed to determine whether they preserve task performance. Consequently, the number of evaluated execution trajectories during pruning can exceed that of the initial candidate set.

Overall, more than 90\% of the wall-clock time is spent on executing and validating the initial and pruned multi-agent workflows, whereas candidate construction and topology-generator optimization contribute only a small fraction of the total cost. This indicates that the curriculum-training overhead is dominated by repeated multi-agent execution rather than computationally expensive parameter optimization.

Table~\ref{tab:curriculum_overall_cost} further summarizes the overall execution and token cost. The token counts include all prompts and completions generated during initial candidate execution, repeated empirical evaluation, pruned-topology re-execution, and KG-Verifier calls. Dense retrieval, graph pruning, connectivity checking, and gradient-based generator optimization do not incur LLM token consumption.

\begin{table}[t]
    \centering
    \caption{\textbf{Overall Curriculum Optimization Cost.}
    The reported token consumption includes all LLM calls involved in execution-based supervision collection and verification.}
    \label{tab:curriculum_overall_cost}
    \resizebox{\columnwidth}{!}{
    \begin{tabular}{l|cccccc}
        \toprule
        \textbf{Dataset} &
        \textbf{Instances} &
        \textbf{Candidates} &
        \textbf{$R$} &
        \textbf{Hardware} &
        \textbf{Tokens} &
        \textbf{Time} \\
        \midrule

        \textsc{AQuA}
        & 40
        & 5
        & 2
        & 1$\times$ NVIDIA A6000
        & $\approx$ 0.52M
        & $\approx$ 4 h \\

        \textsc{GPQA}
        & 40
        & 5
        & 2
        & 1$\times$ NVIDIA A6000
        & $\approx$ 1.11M
        & $\approx$ 5 h \\

        \bottomrule
    \end{tabular}
    }
\end{table}

\section{Additional Experiments with Stronger Backbones}
\label{app:stronger_backbone}

To examine whether the effectiveness of K-GAT is limited to relatively small backbone models, we conduct an additional evaluation with a substantially stronger model.
Specifically, we use \textsc{Qwen3-235B-A22B} as the backbone for all compared methods and evaluate them on \textsc{HLE Bio/Chem Gold}, a challenging knowledge-intensive benchmark containing 149 samples.
This setting allows us to test whether evidence-conditioned topology generation remains beneficial when the underlying language model already has strong reasoning and knowledge capabilities.

Table~\ref{tab:stronger_backbone} reports the results.
K-GAT achieves the best accuracy of 0.1812, outperforming all fixed-topology baselines.
Compared with \textsc{K-GAT (Inf w/o KG)} (0.1342), the full K-GAT also shows a clear improvement, indicating that the retrieved external evidence remains useful even when using a stronger backbone.

\begin{table}[h]
\centering
\small
\caption{Additional evaluation on \textsc{HLE Bio/Chem Gold} using \textsc{Qwen3-235B-A22B} as the backbone.}
\label{tab:stronger_backbone}
\begin{tabular}{lc}
\toprule
Method & Accuracy \\
\midrule
\textsc{Chain} & 0.1140 \\
\textsc{Star} & 0.0939 \\
\textsc{Tree} & 0.1006 \\
\textsc{K-GAT} (w/o KG) & 0.1342 \\
\textsc{K-GAT} & \textbf{0.1812} \\
\bottomrule
\end{tabular}
\end{table}

These results suggest that the gains of K-GAT are not merely caused by compensating for weak backbone capacity.
Even with a substantially stronger model, evidence-conditioned topology generation still provides clear benefits over fixed collaboration structures.
This supports our central claim that the improvement comes from better alignment between retrieved evidence and collaboration structure.
At the same time, we position this experiment as a focused supplementary evaluation on harder evidence-grounded QA, rather than a full evaluation on open-ended agent benchmarks.

\section{Representative Verification Cases}
\label{sec:verification_cases}

We provide two representative cases to illustrate the behavior of the KG-Verifier, including one successful correction and one false rejection.

\begin{table*}[t]
    \centering
    \caption{\textbf{Representative KG-Verifier Cases.} Case A shows a successful correction where the verifier fixes an incorrect upstream answer using retrieved evidence, while Case B shows a false rejection caused by ambiguous or mismatched evidence.}
    \label{tab:representative_verification_cases}
    \resizebox{\textwidth}{!}{
    \begin{tabular}{p{2.2cm}|p{13.6cm}}
        \toprule
        \textbf{Item} & \textbf{Details} \\
        \midrule

        \multicolumn{2}{c}{\textbf{Case A: Successful Correction}} \\
        \midrule

        Question &
        In a genomics pipeline, which errors are particularly difficult to detect because they can silently produce plausible but incorrect results? The candidate errors include incompatible file formats, chromosome-prefix inconsistency, reference-assembly mismatch, incorrect identifier conversion, and ``all of the above.'' \\

        Retrieved Evidence &
        Incompatible file formats and chromosome-prefix inconsistencies typically lead to immediate execution failures or conspicuously empty outputs, whereas reference-assembly mismatches and incorrect identifier conversions may allow the pipeline to complete while silently propagating incorrect results. \\

        Topology &
        $\textsc{Domain Expert} \rightarrow \textsc{KG-Verifier}$ \\

        Node 1 &
        \textbf{Domain Expert:} Selects ``all of the above,'' reasoning that all listed error types can be difficult to detect. \\

        Node 2 &
        \textbf{KG-Verifier:} Identifies the inconsistency between the upstream answer and the retrieved evidence, and revises the answer to ``reference-assembly mismatch and incorrect identifier conversion.'' \\

        Outcome &
        \textbf{Successful correction.} The upstream answer is incorrect, while the verifier-corrected answer matches the ground truth. \\

        \midrule
        \multicolumn{2}{c}{\textbf{Case B: False Rejection}} \\
        \midrule

        Question &
        What is the precession frequency of a spin-$1/2$ particle after the magnetic field is switched from the $z$-direction to the $y$-direction? \\

        Retrieved Evidence &
        The retrieved passage discusses spin evolution and magnetic-field direction, but is ambiguous with respect to the requested frequency and is incorrectly interpreted as introducing an additional geometric projection factor. \\

        Topology &
        $\textsc{Physics Reasoning Agent} \rightarrow \textsc{KG-Verifier}$ \\

        Node 1 &
        \textbf{Physics Reasoning Agent:} Correctly concludes that changing the field direction changes the precession axis but not the magnitude of the Larmor frequency, and therefore answers $\gamma B$. \\

        Node 2 &
        \textbf{KG-Verifier:} Incorrectly interprets the retrieved evidence as requiring a geometric projection factor and revises the answer from $\gamma B$ to $\gamma B/\sqrt{2}$. \\

        Outcome &
        \textbf{False rejection.} The upstream answer is correct, but the verifier changes it to an incorrect answer. \\

        \bottomrule
    \end{tabular}
    }
\end{table*}

These cases illustrate both the corrective benefit of evidence-grounded verification and its dependence on the relevance and correct interpretation of the retrieved evidence.

\section{Knowledge Graph Construction}
\label{app:kg_construction}

K-GAT relies on an external structured knowledge graph constructed from large-scale Wikipedia corpora. 
To build the knowledge source, we follow the ontology-guided structured extraction pipeline introduced in StructSense~\cite{chhetri2025structsense}, which converts unstructured textual documents into provenance-grounded structured evidence units. 
Since the knowledge graph construction procedure itself is not the focus of this work, we directly adopt their framework without additional modification.

Specifically, the pipeline first parses Wikipedia documents into structured text segments and extracts entities and relational evidence using an agent-based information extraction framework. 
The extracted entities are then aligned with ontology concepts through hybrid retrieval and concept mapping, producing normalized knowledge units associated with provenance information, including source documents and supporting text spans. 
Finally, the extracted triplets and provenance metadata are organized into a structured knowledge graph for downstream retrieval and topology generation.

Formally, each retrieved evidence unit is represented as:
\begin{equation}
\tau = (h, r, t, \pi),
\end{equation}
where $h$ and $t$ denote the head and tail entities, $r$ denotes the relation type, and $\pi$ represents the provenance information linked to the original Wikipedia evidence span.

During inference, K-GAT retrieves top-$K$ evidence units from the constructed knowledge graph according to dense embedding similarity and uses the retrieved evidence context to condition collaboration topology generation.

Table~\ref{tab:kg_examples} presents several examples of the constructed knowledge units.

\begin{table}[h]
\centering
\small
\caption{Examples of provenance-grounded knowledge units constructed from Wikipedia corpora.}
\label{tab:kg_examples}
\begin{tabular}{p{1.8cm}p{1.4cm}p{1.8cm}}
\toprule
Head Entity & Relation & Tail Entity \\
\midrule
Albert Einstein & developed & Theory of Relativity \\
CRISPR-Cas9 & used for & Genome Editing \\
Transformer & introduced in & Attention Is All You Need \\
Parkinson's Disease & affects & Dopaminergic Neurons \\
\bottomrule
\end{tabular}
\end{table}

The provenance-linked construction process enables K-GAT to trace intermediate reasoning steps back to concrete Wikipedia evidence spans, improving both evidence grounding and reasoning interpretability.

\section{Detailed Agent Profiles and Prompts}
\label{sec:appendix_profiles}

In this section, we provide the detailed system prompts (role descriptions) and allowable connection graphs used for each dataset.

% ---------------------------------------------------------
\subsection{Mathematical Reasoning (AQUA-RAT \& GSM8K)}
\label{app:profiles_math}

For mathematical reasoning tasks, we assign complementary roles covering symbolic reasoning, numerical computation, and verification. The collaboration structure emphasizes cross-checking between analytical and programmatic reasoning processes.

\begin{table}[t]
\centering
\small
\setlength{\tabcolsep}{4pt}
\renewcommand{\arraystretch}{1.1}
\caption{Agent Role Profiles for AQUA-RAT and GSM8K.}
\label{tab:math_roles}
\begin{tabularx}{\columnwidth}{@{}p{2.7cm}>{\raggedright\arraybackslash}X@{}}
\toprule
\textbf{Role} & \textbf{System Prompt / Instructions} \\
\midrule

\textbf{Math Solver} &
You are a math expert. You will be given a problem and hints from other agents. Provide a step-by-step solution based on the hints.
\textit{AQUA-RAT:} the last line must be \texttt{The answer is X}.
\textit{GSM8K:} the last line must be \texttt{The answer is N}. \\

\textbf{Mathematical Analyst} &
You are a mathematical analyst. First provide an abstract solution template; then substitute concrete values to compute the final result.
End with \texttt{The answer is X} (AQUA-RAT) or \texttt{The answer is N} (GSM8K). \\

\textbf{Programming Expert} &
Integrate reasoning and Python to solve the problem.
\textit{AQUA-RAT:} map the result to (A--E) and output \texttt{The answer is X}.
\textit{GSM8K:} compute a numeric result and output \texttt{The answer is N}. \\

\textbf{Inspector} &
Check whether the logic/calculation is correct and whether the code matches the analysis (if present). Provide corrections if needed.
End with \texttt{The answer is X} / \texttt{The answer is N}. \\

\bottomrule
\end{tabularx}
\end{table}

\paragraph{Topological constraints.}
\begin{itemize}
    \item \textbf{Mathematical Analyst} connects to \textbf{Math Solver}, \textbf{Programming Expert}, and \textbf{Inspector}.
    \item \textbf{Math Solver} and \textbf{Programming Expert} communicate bidirectionally.
    \item \textbf{Inspector} communicates bidirectionally with all reasoning roles.
\end{itemize}

% ---------------------------------------------------------
\subsection{Graduate-Level Science (GPQA)}
\label{app:profiles_gpqa}

For graduate-level scientific reasoning, we employ specialized expert roles together with skeptical verification agents to encourage rigorous multi-step deduction and cross-validation.

\begin{table}[t]
\centering
\small
\setlength{\tabcolsep}{4pt}
\renewcommand{\arraystretch}{1.1}
\caption{Agent Role Profiles for GPQA.}
\label{tab:gpqa_roles}
\begin{tabularx}{\columnwidth}{@{}p{2.9cm}>{\raggedright\arraybackslash}X@{}}
\toprule
\textbf{Role} & \textbf{Role Description} \\
\midrule

\textbf{Domain Knowledge Expert} &
Provide theoretical foundations and assumptions. End with \texttt{The answer is X}. \\

\textbf{Analytical Solver} &
Perform rigorous step-by-step deduction. End with \texttt{The answer is X}. \\

\textbf{Option Eliminator} &
Eliminate incorrect options with justifications. End with \texttt{The answer is X}. \\

\textbf{Critical Reviewer} &
Act as a skeptic; attempt to falsify. End with \texttt{The answer is X}. \\

\bottomrule
\end{tabularx}
\end{table}

\paragraph{Topological constraints.}
\begin{itemize}
    \item All agents connect bidirectionally with the \textbf{Critical Reviewer}.
    \item \textbf{Domain Knowledge Expert} provides information to both \textbf{Analytical Solver} and \textbf{Option Eliminator}.
    \item \textbf{Analytical Solver} and \textbf{Option Eliminator} communicate bidirectionally for cross-verification.
\end{itemize}

% ---------------------------------------------------------
\subsection{Code Generation (HumanEval)}
\label{app:profiles_humaneval}

For code generation tasks, the workflow follows a software-engineering-style collaboration process including planning, implementation, testing, and debugging.

\begin{table}[t]
\centering
\small
\setlength{\tabcolsep}{4pt}
\renewcommand{\arraystretch}{1.1}
\caption{Agent Role Profiles for HumanEval.}
\label{tab:code_roles}
\begin{tabularx}{\columnwidth}{@{}p{2.8cm}>{\raggedright\arraybackslash}X@{}}
\toprule
\textbf{Role} & \textbf{System Prompt} \\
\midrule

\textbf{Project Manager} &
Oversee overall code structure. Suggest optimal design patterns. Keep replies concise (within 50 words). \\

\textbf{Algorithm Designer} &
Specify algorithm design, including functions/classes. Provide pseudocode for complex logic. \\

\textbf{Programming Expert} &
Write the full Python implementation based on the design. Do not change signatures. Output code only. \\

\textbf{Test Analyst} &
Provide edge cases and failure scenarios based on the signature/docstring. \\

\textbf{Bug Fixer} &
Modify and improve the code based on test feedback; output updated code only. \\

\bottomrule
\end{tabularx}
\end{table}

\paragraph{Topological constraints.}
\begin{itemize}
    \item The default execution flow is:
    \textbf{Project Manager}
    $\rightarrow$
    \textbf{Algorithm Designer}
    $\rightarrow$
    \textbf{Programming Expert}
    $\rightarrow$
    \textbf{Test Analyst}
    $\rightarrow$
    \textbf{Bug Fixer}.
    
    \item \textbf{Bug Fixer} can send feedback to \textbf{Programming Expert}.
    
    \item \textbf{Algorithm Designer} communicates with both \textbf{Project Manager} and \textbf{Test Analyst}.
\end{itemize}

% ---------------------------------------------------------
\subsection{Multi-Disciplinary Knowledge (MMLU \& MMLU-Pro)}
\label{app:profiles_mmlu}

For multi-disciplinary knowledge reasoning, we instantiate domain-specific expert roles together with a critic agent for consistency verification.

\begin{equation}
\small
\begin{aligned}
\mathcal{R}=\{
&\text{Math},
\text{CS},
\text{Physics},
\text{Chemistry}, \\
&\text{Biology/Health},
\text{Economics/Business}, \\
&\text{Law},
\text{History},
\text{Engineering},
\text{Critic}
\}.
\end{aligned}
\end{equation}

\begin{table}[t]
\centering
\small
\setlength{\tabcolsep}{4pt}
\renewcommand{\arraystretch}{1.1}
\caption{Role descriptions for MMLU and MMLU-Pro.}
\label{tab:mmlu_roles}
\begin{tabularx}{\columnwidth}{@{}p{2.5cm}>{\raggedright\arraybackslash}X@{}}
\toprule
\textbf{Role} & \textbf{Description} \\
\midrule

Domain Experts &
Analyze step-by-step using domain-specific theories; justify the correct option. \\

Critic &
Verify reasoning, identify fallacies, and confirm the final answer. \\

\bottomrule
\end{tabularx}
\end{table}

\paragraph{Topological constraints.}
\begin{itemize}
    \item Every Domain Expert connects bidirectionally to the \textbf{Critic}.
    \item STEM-related experts communicate with mathematically related domains.
    \item Humanities-related experts communicate with law and economics domains.
    \item Cross-disciplinary communication is allowed between CS, Engineering, and Economics/Business.
\end{itemize}

% ---------------------------------------------------------
\subsection{Reasoning Strategies (StrategyQA)}
\label{app:profiles_strategyqa}

For StrategyQA, the collaboration workflow focuses on decomposition, contextual grounding, and causal reasoning.

\begin{table}[t]
\centering
\small
\setlength{\tabcolsep}{4pt}
\renewcommand{\arraystretch}{1.1}
\caption{Agent Role Profiles for StrategyQA.}
\label{tab:strat_roles}
\begin{tabularx}{\columnwidth}{@{}p{2.9cm}>{\raggedright\arraybackslash}X@{}}
\toprule
\textbf{Role} & \textbf{Functionality} \\
\midrule

\textbf{Decomposition Strategist} &
Breaks the main question into 2--3 sub-questions and synthesizes answers for a final True/False decision. \\

\textbf{Context Specialist} &
Identifies key entities and provides background context to bridge implicit gaps. \\

\textbf{Logical Reasoner} &
Performs linear step-by-step deduction from premises using cause-and-effect reasoning. \\

\textbf{Critical Reviewer} &
Scrutinizes reasoning for fallacies; derives conclusion independently if needed. \\

\bottomrule
\end{tabularx}
\end{table}

\paragraph{Topological constraints.}
\begin{itemize}
    \item \textbf{Critical Reviewer} communicates bidirectionally with all other roles.
    \item \textbf{Decomposition Strategist} communicates with both \textbf{Context Specialist} and \textbf{Logical Reasoner}.
    \item \textbf{Context Specialist} and \textbf{Logical Reasoner} exchange grounding information bidirectionally.
\end{itemize}

\end{document}